\documentclass{article}

\usepackage{microtype}
\usepackage{graphicx}
\usepackage{stfloats}
\usepackage{booktabs}
\usepackage{multirow}
\usepackage{hyperref}
\usepackage{amsmath}
\usepackage{algorithm}
\usepackage{listings}
\usepackage{icml2025}
\makeatletter
\gdef\isaccepted{1}
\renewcommand{\Notice@String}{}
\makeatother

\icmltitlerunning{Beyond Rows to Reasoning: Agentic Retrieval for Multimodal Spreadsheet Understanding and Editing}

\begin{document}

\twocolumn[
\icmltitle{\smash{\makebox[0pt][r]{\raisebox{-2.2em}{\includegraphics[height=4em]{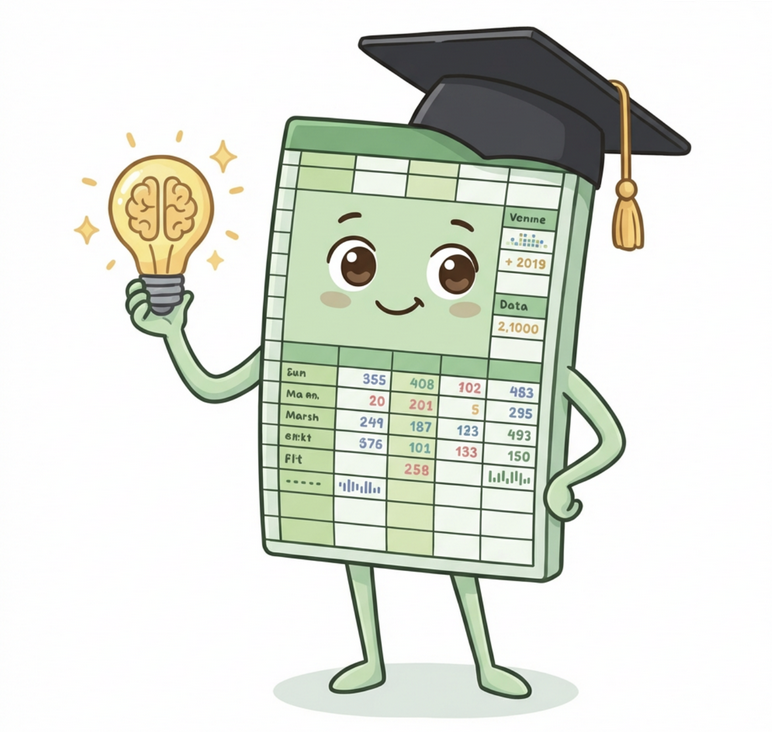}}\hspace{0.5em}}}Beyond Rows to Reasoning: Agentic Retrieval for \\ Multimodal Spreadsheet Understanding and Editing}

\icmlsetsymbol{equal}{*}

\begin{icmlauthorlist}
\icmlauthor{Anmol Gulati}{pwc}
\icmlauthor{Sahil Sen}{pwc}
\icmlauthor{Waqar Sarguroh}{pwc}
\icmlauthor{Kevin Paul}{pwc}
\end{icmlauthorlist}

\icmlaffiliation{pwc}{Commercial Technology and Innovation Office, PricewaterhouseCoopers, U.S.}
\icmlcorrespondingauthor{Anmol Gulati}{anmol.gulati@pwc.com}

\icmlkeywords{Multimodal RAG, Spreadsheet understanding, Agentic retrieval, Enterprise QA, Tool-calling agents}

\vskip 0.3in
]

\printAffiliationsAndNotice{}

\begin{figure*}[!b]
    \centering
    \includegraphics[width=0.88\textwidth]{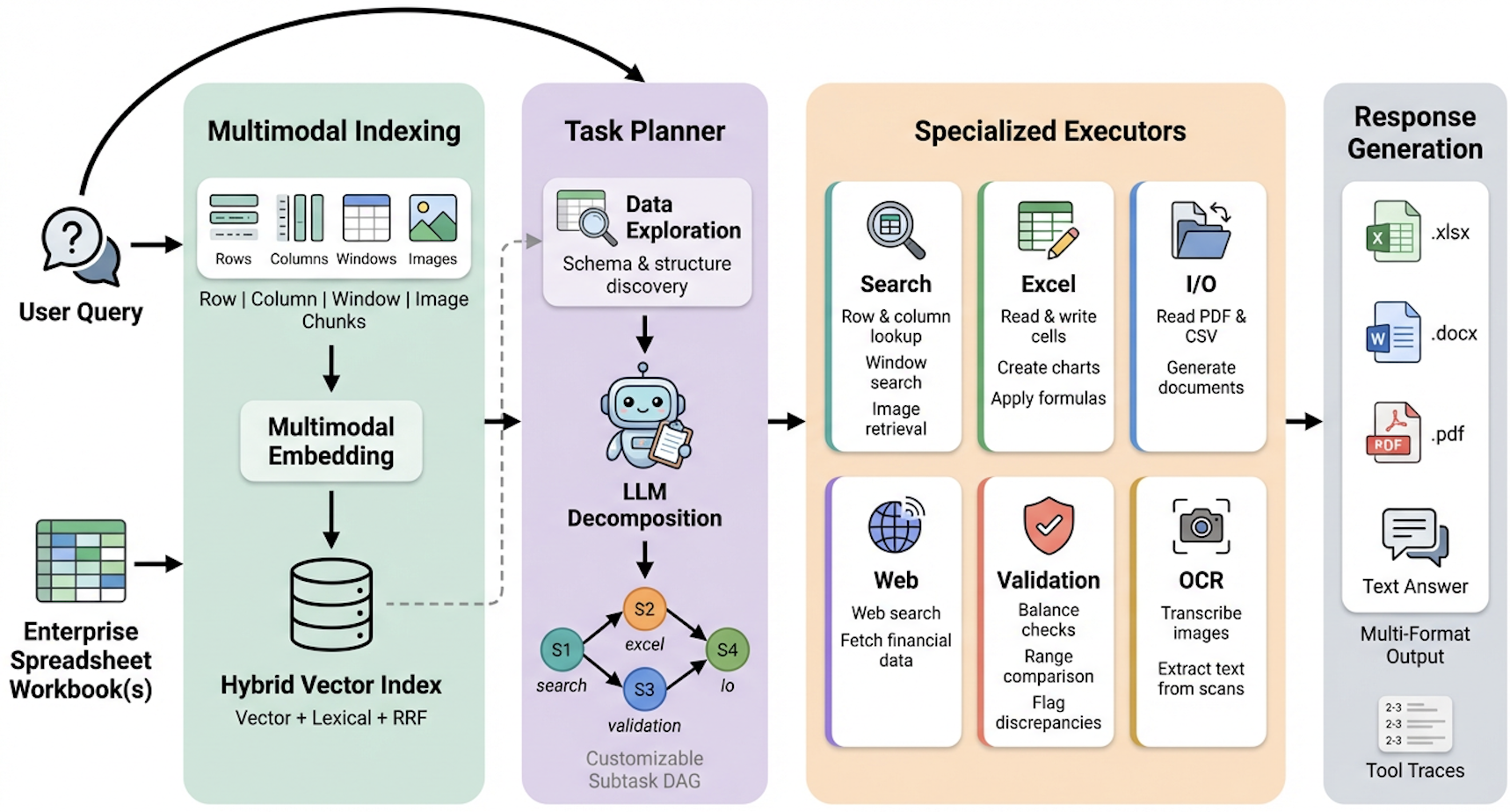}
    \caption{Overview of the BRTR framework pipeline: multimodal spreadsheet indexing, agentic task planning and decomposition, specialized tool execution, and multi-format response generation with full end-to-end tool-trace auditability.}

    \label{fig:architecture}
\end{figure*}

\begin{abstract}

Recent advances in multimodal Retrieval-Augmented Generation (RAG) enable Large Language Models (LLMs) to analyze enterprise spreadsheet workbooks containing millions of cells, cross-sheet dependencies, and embedded visual artifacts. However, state-of-the-art approaches exclude critical context through single-pass retrieval, lose data resolution through compression, and exceed LLM context windows through naive full-context injection, preventing reliable multi-step reasoning over complex enterprise workbooks. We introduce Beyond Rows to Reasoning (BRTR), a multimodal agentic framework for spreadsheet understanding that replaces single-pass retrieval with an iterative tool-calling loop, supporting end-to-end Excel workflows from complex analysis to structured editing. Supported by over 200 hours of expert human evaluation, BRTR achieves state-of-the-art performance across three frontier spreadsheet understanding benchmarks, surpassing prior methods by 25 percentage points on FRTR-Bench, 7 points on SpreadsheetLLM, and 32 points on FINCH. We evaluate five multimodal embedding models, identifying NVIDIA NeMo Retriever 1B as the top performer for mixed tabular and visual data, and vary nine LLMs. Ablation experiments confirm that the planner, retrieval, and iterative reasoning each contribute substantially, and cost analysis shows GPT-5.2 achieves the best efficiency--accuracy trade-off. Throughout all evaluations, BRTR maintains full auditability through explicit tool-call traces.
\end{abstract}

\begin{figure*}[t]
    \centering
    \includegraphics[width=0.91\textwidth]{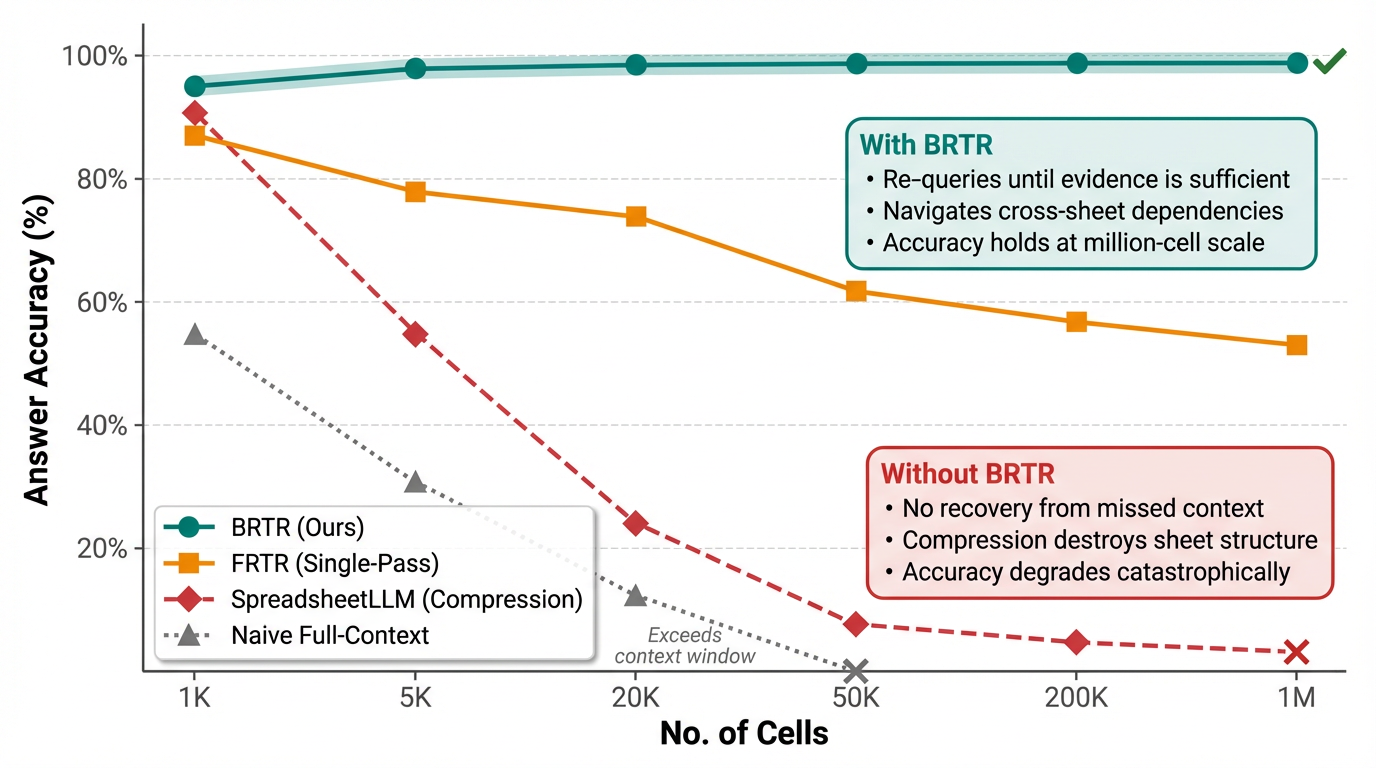}
    \caption{Conceptual illustration of answer accuracy as a function of spreadsheet size. BRTR maintains near-perfect accuracy across all scales by iteratively re-querying until evidence is sufficient, while single-pass and compression-based methods degrade as workbook complexity increases. Na\"ive full-context approaches exceed LLM context windows beyond 50K cells. Trends are derived from aggregate observations across evaluations; see Tables~\ref{tab:frtr_bench_results} and~\ref{tab:spreadsheetllm_results} for precise measurements.}
    \label{fig:main_graph}
\end{figure*}

\section{Introduction}

Recent advances in multimodal Retrieval-Augmented Generation (RAG) \citep{ref_rag} have enabled Large Language Models (LLMs) to analyze enterprise spreadsheet workbooks containing complex tabular data, cross-sheet dependencies, and embedded visual artifacts \citep{ref_spreadsheetllm, ref_frtr}. Current approaches follow two main strategies: compression-based methods encode worksheets into condensed representations that fit within LLM context windows \citep{ref_spreadsheetllm}, while retrieval-based methods chunk and embed spreadsheet content for semantic search at query time \citep{ref_frtr}. However, both commit to a single pass with no mechanism for iterative refinement. Real analysts cross-reference across sheets, follow cell dependencies, and gather evidence incrementally before drawing conclusions; current pipelines cannot mimic this behavior because the LLM lacks the ability to re-query, explore adjacent regions, or refine its search based on intermediate results. Additionally, existing work has not comprehensively evaluated multimodal embedding models for spreadsheet content, leaving practitioners without guidance on which embeddings best capture the semantic structure of mixed tabular and visual data.

In this paper, we introduce Beyond Rows to Reasoning (BRTR), a multimodal agentic framework for spreadsheet understanding that replaces single-pass retrieval with an iterative tool-calling loop, enabling LLMs to invoke structured search tools, inspect results, and refine queries across multiple turns for end-to-end Excel analysis and editing (Figure~\ref{fig:architecture}). We make two contributions:
\begin{itemize}
    \item \textbf{Multimodal embedding model evaluation:} We present a comprehensive comparison of five multimodal embedding models for spreadsheet content retrieval on FRTR-Bench, providing practitioners with the first empirical guidance on embedding selection for mixed tabular and visual data (Section~\ref{subsec:embedding_eval}).
    \item \textbf{BRTR agentic framework:} BRTR equips LLMs with structured search tools and a context budgeting mechanism that controls token growth. Supported by +200 hours of human evaluation, BRTR achieves state-of-the-art accuracy on FRTR-Bench and the SpreadsheetLLM benchmark, substantially outperforming single-pass retrieval and compression baselines. We further extend BRTR with spreadsheet manipulation and multimodal interaction tools, achieving state-of-the-art performance on the FINCH finance and accounting benchmark and demonstrating generalization to composite, multi-step enterprise workflows spanning calculation, formatting, visualization, and cross-file validation across heterogeneous artifacts including spreadsheets, PDFs, and images.
\end{itemize}

\section{Related Work}

\subsection{Retrieval-based Spreadsheet Understanding}

Spreadsheet understanding poses unique challenges due to two-dimensional layouts, flexible formatting, and mixed content types. Early table understanding methods adapted transformer architectures for structured data. TaPas \citep{ref_tapas} extended BERT to encode tables by flattening rows and adding positional embeddings for columns and rows, enabling weakly supervised question answering without generating logical forms. TUTA \citep{ref_tuta} introduced tree-based transformers with a bi-dimensional coordinate tree to capture both spatial and hierarchical structure, achieving state-of-the-art performance on cell type and table type classification tasks through structure-aware attention mechanisms.

For enterprise spreadsheets, compression and retrieval approaches have emerged to handle workbooks that exceed LLM context limits. SpreadsheetLLM \citep{ref_spreadsheetllm} introduced SheetCompressor, an encoding framework with structural-anchor-based compression, inverse index translation, and data-format-aware aggregation, achieving 25$\times$ compression while improving table detection by 25.6\%. From Rows to Reasoning (FRTR) \citep{ref_frtr} proposed multi-granular chunking at the row, column, and window levels with hybrid lexical-dense retrieval, enabling question answering over workbooks with millions of cells. Upstream of retrieval, LlamaSheets \citep{ref_llamasheets} addresses the pre-processing stage, using hierarchical header extraction and adaptive table segmentation to transform messy spreadsheets into structured data for downstream consumption. While these methods advance scalability, they commit to a single pass: the system retrieves or extracts a fixed context before generation, with no mechanism for iterative refinement if initial retrieval misses critical evidence.

\subsection{Agentic and Iterative Retrieval for Structured Data}

Recent work explores multi-step interaction and tool-based exploration for table understanding. These agentic approaches build on the ReAct paradigm \citep{ref_react}, which demonstrated that interleaving reasoning traces with actions enables LLMs to recover from errors and update plans based on external feedback, and on broader advances in agentic retrieval-augmented generation and tool selection \citep{ref_agenticrag, ref_toolagentselection}. Several systems adopt code execution as their primary action space. SheetAgent \citep{ref_sheetagent} introduces a three-module architecture (Planner, Informer, Retriever) where an LLM generates Python code to manipulate spreadsheets, with SQL-based subview generation to narrow reasoning scope, achieving 20--40\% improvements on spreadsheet manipulation benchmarks. TableZoomer \citep{ref_tablezoomer} proposes query-aware table zooming to dynamically generate sub-table schemas through column selection and entity linking, combined with Program-of-Thoughts execution to reduce numerical hallucination. SheetBrain \citep{ref_sheetbrain} adopts a neuro-symbolic dual-workflow design where a Python sandbox with an Excel helper toolkit handles multi-turn execution, and a validation module triggers re-execution when answers fail correctness checks.

Other agentic methods replace free-form code generation with structured command grammars or learned policies. SheetMind \citep{ref_sheetmind} deploys a multi-agent framework as a Google Sheets extension, with a Manager agent that decomposes instructions, an Action agent that generates structured commands via a Backus--Naur Form grammar, and a Reflection agent that validates alignment with user intent, achieving approximately 80\% success on single-step tasks and 70\% on multi-step instructions. TableMind \citep{ref_tablemind} trains an autonomous programmatic agent through supervised fine-tuning followed by reinforcement learning with a rank-aware policy optimization algorithm, enabling multi-turn tool invocation with a plan-action-reflect cycle. However, these methods either depend on code execution, which can be unreliable when agents enter unproductive loops, or require model fine-tuning with specialized supervision. BRTR avoids both limitations by equipping general-purpose LLMs with structured search tools through native function-calling, requiring no fine-tuning and constraining tool calls to prevent runaway loops.

\subsection{Model-Centric Table Reasoning}

An alternative line of work develops specialized pre-training objectives for table reasoning. TAPEX \citep{ref_tapex} pre-trains language models to mimic SQL execution over synthetic queries, ForTaP \citep{ref_fortap} leverages spreadsheet formulas for numerical reasoning through structure-aware pre-training, and Fortune \citep{ref_fortune} applies reinforcement learning to formula generation. These approaches require substantial training resources and specialized supervision (execution traces, formula annotations, or RL infrastructure), limiting accessibility for enterprise deployment. BRTR instead leverages the native reasoning and function-calling capabilities of general-purpose LLMs, requiring no task-specific training.

\subsection{LLM-Powered Spreadsheet Products}

A growing number of commercial products embed LLMs directly into spreadsheet applications. Claude for Excel \citep{ref_claudeexcel}, Microsoft Copilot Agent Mode \citep{ref_copilotagent}, Gemini in Google Sheets \citep{ref_geminisheets}, and Claude Cowork \citep{ref_claudecowork} each provide varying degrees of in-workbook AI assistance, from formula generation to agentic planning, but all remain scoped to the active workbook or process files on demand without semantic indexing. Recent benchmarks including SpreadsheetBench \citep{ref_spreadsheetbench} (912 tasks from online forums), InstructExcel \citep{ref_instructexcel} (natural language instruction in Excel), FINCH \citep{ref_finch} (172 enterprise workflows across 1{,}710 spreadsheets), and SpreadsheetArena \citep{ref_spreadsheetarena} (arena-style preference evaluation of end-to-end workbook generation across 16 models) have standardized evaluation of these systems. BRTR differs by indexing content into a searchable vector store, enabling agentic cross-file retrieval over thousands of enterprise documents with full tool-trace auditability.

\section{Methodology}\label{sec:methodology}

BRTR utilizes a retrieval pipeline with an agentic loop that enables iterative evidence acquisition. The framework consists of three core components: (1) multimodal indexing and retrieval, which chunks spreadsheet content and embeds it for semantic search (Section~\ref{subsec:indexing}); (2) an agentic tool-calling loop, where the LLM invokes structured search tools and refines queries across multiple turns (Section~\ref{subsec:agentic_loop}); and (3) a planner-executor architecture that decomposes composite workflows into dependency graphs of sub-tasks dispatched to specialized executors (Section~\ref{subsec:planner_executor}). Figure~\ref{fig:architecture} illustrates the complete pipeline.

\subsection{Multimodal Indexing and Retrieval}\label{subsec:indexing}

BRTR builds on the FRTR chunking and retrieval infrastructure~\citep{ref_frtr} (Figure~\ref{fig:architecture}, left). Given a workbook, the system extracts four chunk types (row, column, rectangular window, and embedded image), each preserving metadata such as sheet name, cell coordinates, and headers. All chunks are embedded using a multimodal embedding model; based on the evaluation in Section~\ref{subsec:embedding_eval}, we adopt NVIDIA NeMo Retriever 1B.

At query time, BRTR performs hybrid retrieval combining dense cosine-similarity search with BM25 lexical matching, the latter being important for exact numerical values and cell references. Results across all chunk types are fused via Reciprocal Rank Fusion (RRF)~\citep{ref_rrf}:
\begin{equation}\label{eq:rrf}
\text{score}_{\text{RRF}}(c, q) = \sum_{r \in \mathcal{R}} \frac{1}{k + \text{rank}_r(c, q)}
\end{equation}
where $\mathcal{R}$ is the set of ranked lists (dense, lexical $\times$ chunk type), $\text{rank}_r(c, q)$ is the position of chunk $c$ in list $r$, and $k = 60$. The top-$K$ chunks ($K = 10$) are passed to the agent as initial context before iterative refinement begins.

\begin{algorithm}[t]
\caption{BRTR Planner-Executor Pipeline}\label{alg:brtr_planner}
\begin{algorithmic}[1]
\REQUIRE Task $q$, corpus $\mathcal{C}$, LLM $\mathcal{M}$, $K{=}10$, $T{=}50$
\ENSURE Answer $a$, tool trace $\tau$

\STATE \textbf{Phase 1: Decompose}
\STATE $e \gets \operatorname{ExploreData}(q, \mathcal{C}, K)$ \COMMENT{Vector-search quoted terms for column locations}
\STATE $P, o \gets \mathcal{M}(q, e)$ \COMMENT{Subtask DAG $\{(s_i, \mathit{type}_i, \mathit{deps}_i)\}_{i=1}^{n}$ and output type $o$}

\STATE \textbf{Phase 2: Execute subtask DAG}
\STATE $\mathit{done} \gets \{\};\; \tau \gets [\,]$
\WHILE{$|\mathit{done}| < n$}
    \STATE $\mathit{ready} \gets \{s_i \in P : \mathit{deps}_i \subseteq \mathit{done}\}$
    \FORALL{$s_i \in \mathit{ready}$ \textbf{in parallel}}
        \STATE $\mathit{tools}_i \gets \operatorname{GetToolSet}(\mathit{type}_i)$ \COMMENT{Type-specific tools}
        \STATE $\mathit{ctx}_i \gets \{s_i.\mathit{desc},\; q,\; \operatorname{Summarize}(\mathit{done}[\mathit{deps}_i])\}$
        \FOR{$t = 1, \ldots, T$}
            \STATE $\mathit{resp} \gets \mathcal{M}(\mathit{ctx}_i,\; \mathit{tools}_i)$
            \FORALL{tool call $f$ in $\mathit{resp}$}
                \STATE $r \gets \operatorname{Execute}(f)$
                \STATE $\tau \gets \tau \cup \{(f, r)\};\; \mathit{ctx}_i \gets \mathit{ctx}_i \cup \{f, r\}$
            \ENDFOR
        \ENDFOR
        \STATE $\mathit{done}[s_i] \gets \mathit{resp}$
    \ENDFOR
\ENDWHILE

\STATE \textbf{Phase 3: Synthesize}
\STATE $a \gets \mathcal{M}\!\bigl(q,\; \operatorname{Summarize}(\mathit{done}),\; o\bigr)$ \COMMENT{Merge all executor outputs}
\STATE \textbf{return} $(a,\; \tau)$
\end{algorithmic}
\end{algorithm}

\subsection{Agentic Tool-Calling Loop}\label{subsec:agentic_loop}

The core innovation of BRTR is replacing single-pass retrieval with an iterative tool-calling loop where the LLM invokes search tools, inspects results, and refines queries until sufficient evidence is gathered.
\subsubsection{Search Tools}\label{subsubsec:search_tools}

The agent has access to five search tools: \texttt{search\_rows} retrieves row chunks, \texttt{search\_columns} retrieves column chunks, \texttt{search\_windows} retrieves rectangular cell regions, \texttt{search\_images} retrieves embedded visual content, and \texttt{search\_all} performs a unified search across all content types. Each tool accepts a natural language query and returns the top-$K$ matching chunks with their metadata. All tools use the same hybrid dense-lexical retrieval pipeline described in Section~\ref{subsec:indexing}. Row and column searches additionally accept optional coordinate parameters (row number, column number) that filter results to chunks at specific positions, enabling the agent to combine semantic relevance with precise cell-level targeting.

\subsubsection{Iterative Refinement Process}\label{subsubsec:iterative_process}

Given a user query, the LLM analyzes it to determine which content types are relevant and invokes one or more search tools. After inspecting the returned chunks, the model decides whether additional evidence is needed. If initial results are insufficient or ambiguous, the LLM refines its search query, requests different content types, or applies coordinate filtering to examine specific rows or columns identified in prior results. Each agent invocation is bounded by $T = 50$ tool-call iterations, a limit that prevents unproductive search loops; the loop terminates early when the model determines it has gathered sufficient evidence. In the planner configuration (Section~\ref{subsec:planner_executor}), this budget applies independently to each executor sub-task. In practice, frontier models average 3--6 tool calls per query, well below this ceiling (Appendix~\ref{app:tool_trace} illustrates a representative trace).

\subsubsection{Context Management}\label{subsubsec:context_mgmt}

As the agent accumulates tool-call results across iterations, the conversation context grows. Image chunks, embedded as base64-encoded content, are the primary driver of token consumption. To prevent context overflow, BRTR applies an always-prune strategy: whenever new images are retrieved, all prior image content is stripped from earlier tool-call responses, retaining their textual metadata (sheet name, coordinates, alt-text) while removing the encoded image data. The agent's reasoning about prior images is preserved in the conversation history, so retaining only the most recent batch avoids overflow without sacrificing accumulated understanding.

\subsection{Planner-Executor Architecture}\label{subsec:planner_executor}

To generalize to composite, multi-step enterprise workflows, we extend BRTR with a planner and six specialized executors (Figure~\ref{fig:architecture}). The \emph{planner} decomposes workflow instructions into a dependency graph of sub-tasks (Appendix~\ref{app:decomposition_prompt} provides the prompt template); independent branches execute concurrently. Six executors handle distinct capabilities: \emph{Excel} (read/write cells, formulas, formatting, charts), \emph{IO} (PDFs, CSVs, DOCX), \emph{web} (search, URL fetching, financial data), \emph{validation} (accounting invariants such as balance-sheet identity and debit-credit equality), \emph{OCR} (image transcription and text extraction from scanned documents), and \emph{search} (the base BRTR retrieval tools).

Each executor operates within the agentic loop described above. The planner dispatches ready sub-tasks to the appropriate executor and feeds completed results into downstream dependencies. This decomposition directly addresses the long-horizon composition bottleneck identified in prior work~\citep{ref_finch}, where monolithic agents suffer error accumulation across multi-task workflows.

\section{Experiments}

We evaluate BRTR through four experiments. First, we compare multimodal embedding models to identify the best retrieval backbone for spreadsheet content. Second, we evaluate the full BRTR agentic framework on two benchmarks against existing approaches. Third, we assess BRTR on the FINCH benchmark for generalization. Fourth, we ablate key components to isolate their individual contributions.

\subsection{Experiment 1: Multimodal Embedding Model Comparison}\label{subsec:embedding_eval}

\paragraph{Experimental Setup.}
Five multimodal embedding models are evaluated on a subset of FRTR-Bench comprising 146 queries across 28 files. Each query targets specific spreadsheet content requiring retrieval of relevant chunks containing rows, columns, cell windows, or embedded images. The models span proprietary and open-source architectures across multiple providers: Amazon Titan Multimodal, Cohere Embed v4, Amazon Nova Multimodal, NVIDIA NeMo Retriever 1B, and Jina Embeddings v4. Because spreadsheet data admits many valid retrieval paths to a correct answer, computing exhaustive relevance judgments for each query is infeasible. Instead, we retrieve the top-10 chunks by cosine similarity and evaluate relevance using GPT-5 as an LLM judge, comparing retrieved content against ground-truth annotations. We validate this automated assessment through the downstream human evaluation in Section~\ref{subsec:agentic_eval}: retrieval quality is ultimately reflected in end-to-end answer accuracy, which is judged by human evaluators across all benchmarks. Furthermore, better retrieval can lead to less total searches, reducing the LLM's overall token usage.

\paragraph{Metrics.}
We report standard retrieval metrics at cutoffs $K \in \{5, 10\}$: nDCG@$K$ (normalized Discounted Cumulative Gain) evaluates ranking quality with position-weighted relevance; Recall@$K$ measures the fraction of relevant chunks appearing within the top-$K$ results; and MAP@$K$ (Mean Average Precision) captures precision across all recall levels.

\begin{table}[t]
\centering
\caption{Multimodal embedding model comparison on FRTR-Bench (146 queries, 28 files). Bold indicates best per metric.}
\label{tab:embedding_results}
\resizebox{\columnwidth}{!}{%
\begin{tabular}{lcccccc}
\toprule
& \multicolumn{3}{c}{$K = 5$} & \multicolumn{3}{c}{$K = 10$} \\
\cmidrule(lr){2-4} \cmidrule(lr){5-7}
Embedding Model & nDCG & Recall & MAP & nDCG & Recall & MAP \\
\midrule
Amazon Titan & 0.26 & 0.39 & 0.27 & 0.33 & 0.47 & 0.27 \\
Cohere v4 & \textbf{0.33} & 0.45 & \textbf{0.35} & 0.40 & 0.51 & 0.34 \\
Amazon Nova & 0.30 & 0.45 & 0.32 & 0.40 & 0.58 & 0.33 \\
NVIDIA NeMo & 0.33 & \textbf{0.47} & 0.33 & \textbf{0.42} & \textbf{0.60} & \textbf{0.34} \\
Jina v4 & 0.19 & 0.26 & 0.19 & 0.23 & 0.33 & 0.19 \\
\bottomrule
\end{tabular}%
}
\end{table}

\paragraph{Results.}
Table~\ref{tab:embedding_results} presents retrieval performance across all five embedding models at $K{=}5$ and $K{=}10$. NVIDIA NeMo Retriever achieves the highest Recall@10 (0.60) and also leads in nDCG@10 (0.42) and MAP@10 (0.34), demonstrating the strongest overall retrieval quality. Cohere v4 excels at early-stage ranking, achieving the best nDCG@5 (0.33) and MAP@5 (0.35), suggesting it places relevant results higher, a property that is valuable when retrieving fewer chunks to minimize token consumption. Amazon Nova achieves competitive Recall@10 (0.58) and nDCG@10 (0.40), trailing NVIDIA NeMo by only 2 percentage points on Recall@10. Jina v4 shows substantially lower performance across all metrics, with Recall@10 of only 0.33, a gap of 27 percentage points compared to NVIDIA NeMo, indicating that not all multimodal embedding models generalize effectively to structured spreadsheet content with mixed text and visual elements.

\begin{table*}[t]
\centering
\caption{Cross-technique performance on FRTR-Bench (159 queries, 30 files). BRTR achieves up to 99\% accuracy, 25 percentage points above single-pass FRTR. FRTR and SpreadsheetLLM use mean context sizes of 7{,}691 and 12{,}745 tokens respectively. Claude in Excel supports only Claude models. Operational cost columns report means. Bold indicates highest accuracy per technique.}
\label{tab:frtr_bench_results}
\small
\begin{tabular}{@{}l cccc ccc@{}}
\toprule
& \multicolumn{4}{c}{\textbf{Accuracy}} & \multicolumn{3}{c}{\textbf{BRTR Operational Cost}} \\
\cmidrule(lr){2-5} \cmidrule(lr){6-8}
Model & BRTR & FRTR & SpreadsheetLLM & Claude in Excel & Tool Calls & Tokens & Latency (s) \\
\midrule
GPT-5               & \underline{\textbf{0.99}} & 0.73          & 0.18          & --   & 4.01 & 68,061 & 166.34 \\
Gemini 3 Pro        & \underline{\textbf{0.99}} & 0.68          & 0.29          & --   & 5.40 & 48,462 & 217.77 \\
Claude Opus 4.6     & 0.98          & \textbf{0.74} & \textbf{0.34} & \textbf{0.84} & 5.20 & 462,614 & 143.23 \\
GPT-5.2             & 0.98          & 0.72          & 0.32          & --   & 3.05 & 20,127 & 90.81 \\
Claude Sonnet 4.5   & 0.96          & \textbf{0.74} & 0.13          & 0.69 & 6.20 & 92,202 & 163.83 \\
Gemini 2.5 Pro      & 0.85          & 0.67          & 0.24          & --   & 2.23 & 15,729 & 111.04 \\
GPT-4o              & 0.67          & 0.49          & 0.06          & --   & 3.25 & 50,745 & 140.20 \\
LLaMA 4 Maverick 17B & 0.67         & 0.56          & 0.23          & --   & 1.99 & 38,483 & 92.32 \\
Pixtral Large 2502  & 0.63          & 0.47          & 0.21          & --   & 3.12 & 70,618 & 51.76 \\
\bottomrule
\end{tabular}
\end{table*}

\begin{figure*}[b]
    \centering
    \includegraphics[width=\textwidth]{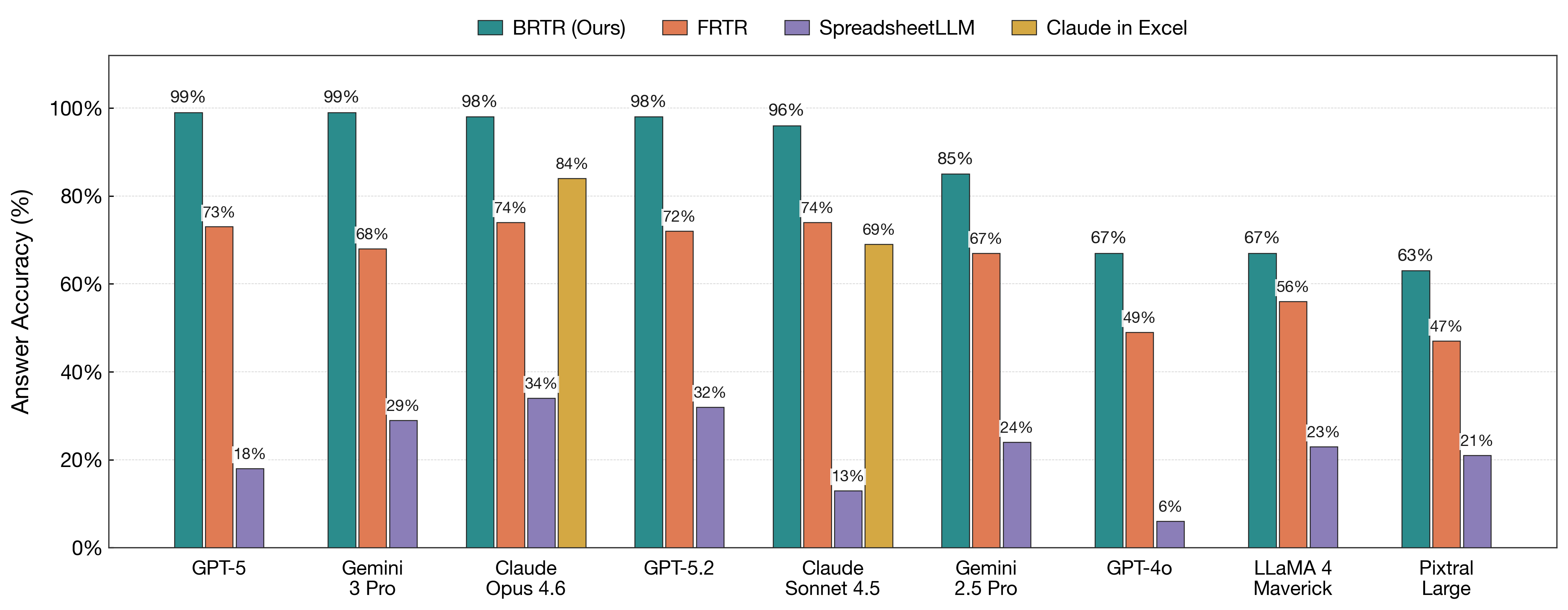}
    \caption{Answer accuracy on FRTR-Bench across nine LLMs and four techniques. BRTR (teal) consistently dominates, achieving up to 99\% accuracy with frontier models, 25 percentage points above the best single-pass FRTR baseline. SpreadsheetLLM compression (lavender) performs poorly throughout, confirming that cross-sheet references require retrieval rather than compression alone.}
    \label{fig:frtr_bar}
\end{figure*}

\paragraph{Discussion.}
For agentic systems like BRTR that iteratively refine searches, higher Recall@10 is preferable, making breadth of initial retrieval more valuable than precision at shallow cutoffs. Based on these findings, we adopt NVIDIA NeMo Retriever as the default embedding model for all subsequent experiments.

\subsection{Experiment 2: BRTR Agentic Framework Evaluation}\label{subsec:agentic_eval}

\paragraph{Experimental Setup.}
BRTR is evaluated across nine LLMs spanning frontier and open-weight models, using NVIDIA NeMo Retriever embeddings based on Experiment~1 findings. Two baselines are compared on both benchmarks: (1) FRTR~\citep{ref_frtr}, single-pass hybrid retrieval followed by generation; and (2) SpreadsheetLLM~\citep{ref_spreadsheetllm}, SheetCompressor encoding of entire workbooks. On FRTR-Bench we additionally report Claude in Excel~\citep{ref_claudeexcel} as a commercial product baseline; on the SpreadsheetLLM benchmark we additionally include na\"ive full-context, which passes uncompressed spreadsheet content directly to the LLM. Claude in Excel is evaluated only on FRTR-Bench: the SpreadsheetLLM benchmark is already well-saturated by existing methods, so we prioritized the more challenging FRTR-Bench. It is excluded from FINCH because it operates within a single workbook and cannot ingest multiple spreadsheets to produce new output files, disqualifying it from the majority of FINCH workflows.

The evaluation spans two benchmarks. FRTR-Bench consists of 159 queries across 30 enterprise workbooks spanning 155 sheets, nearly four million cells, 656K rows, and 53 embedded images~\citep{ref_frtr}, where each query requires locating specific cells, cross-referencing multiple sheets, or interpreting embedded charts. The SpreadsheetLLM benchmark comprises 298 single-sheet, text-only questions requiring numerical reasoning, formula synthesis, cell lookup, and format interpretation across diverse individual worksheets. All BRTR configurations use top-$K{=}10$ initial retrieval with a maximum of 50 tool calls per query. We do not compare against agentic systems such as SheetAgent~\citep{ref_sheetagent}, SheetBrain~\citep{ref_sheetbrain}, or TableMind~\citep{ref_tablemind} because they target different task formulations, relying on Python code execution or requiring supervised fine-tuning and reinforcement learning, making direct comparison on retrieval-centric benchmarks methodologically unsound; additionally, several lack publicly available implementations or evaluate on benchmarks with no overlap with ours.

\paragraph{Metrics.}
Answer Accuracy measures the fraction of queries answered correctly, with correctness determined by semantic equivalence to ground truth, judged by a team of evaluators with hands-on experience in enterprise finance and accounting data, assisted by LLM-based pre-screening but with final correctness decisions made by human reviewers. Evaluators compared each model response against ground-truth annotations across all model and technique configurations on FRTR-Bench (159 queries), the SpreadsheetLLM benchmark (298 queries), and the FINCH benchmark (172 workflows), totaling over 200 person-hours. Mean Tool Calls captures the average number of search tool invocations per query. Mean Tokens reports total token consumption across input context, tool results, and generated responses. Mean Latency measures end-to-end inference time from query submission to final answer.

\begin{table*}[t]
\centering
\caption{Cross-technique performance on the SpreadsheetLLM benchmark (298 queries). BRTR achieves 97--98\% accuracy with frontier models, while SpreadsheetLLM compression reaches 89--91\% and na\"ive full-context approaches yield only 55--68\%. FRTR, SpreadsheetLLM, and Na\"ive Full-Context use mean context sizes of 6{,}920, 5{,}811, and 13{,}631 tokens respectively. Operational cost columns report means. Bold indicates highest accuracy per technique. Models sorted by BRTR accuracy.}
\label{tab:spreadsheetllm_results}
\small
\begin{tabular}{@{}l cccc cc@{}}
\toprule
& \multicolumn{4}{c}{\textbf{Accuracy}} & \multicolumn{2}{c}{\textbf{BRTR Cost}} \\
\cmidrule(lr){2-5} \cmidrule(lr){6-7}
Model & BRTR & FRTR & SpreadsheetLLM & Na\"ive Full-Context & Tokens & Latency (s) \\
\midrule
GPT-5                & \underline{\textbf{0.98}} & \textbf{0.87} & 0.90          & \textbf{0.68} & 156,306 & 273.36 \\
Claude Sonnet 4.5    & 0.97          & 0.84          & \textbf{0.91} & 0.55          & 166,117 & 280.59 \\
Gemini 2.5 Pro       & 0.79          & 0.85          & 0.89          & 0.67          & 27,675  & 223.79 \\
GPT-4o               & 0.56          & 0.77          & 0.78          & 0.35          & 256,523 & 315.93 \\
LLaMA 4 Maverick 17B & 0.30          & 0.53          & 0.73          & 0.41          & 18,608  & 408.73 \\
\bottomrule
\end{tabular}
\end{table*}

\paragraph{Results on FRTR-Bench.}
Table~\ref{tab:frtr_bench_results} and Figure~\ref{fig:frtr_bar} present performance across all approaches on FRTR-Bench. BRTR with frontier models achieves exceptional accuracy: GPT-5 and Gemini 3 Pro both reach 99\% (157/159 queries correct), an improvement of 25 percentage points over the best single-pass FRTR baseline (Claude Opus 4.6 and Claude Sonnet 4.5, both at 74\%). Claude Opus 4.6 and GPT-5.2 each achieve 98\% with BRTR; among these, GPT-5.2 demonstrates notable efficiency with only 20,127 mean tokens and 90.81s mean latency, the lowest resource consumption among high-accuracy configurations. Claude Sonnet 4.5 reaches 96\% accuracy but requires 6.20 tool calls on average compared to 3.05 for GPT-5.2, suggesting a more exploratory search strategy. In addition, while Claude Opus 4.6 approaches state-of-the-art performance, it consumes nearly 10x the amount of tokens of gemini 3 pro, showing a tradeoff between the length of thinking and the answer accuracy.

Iterative retrieval benefits models across capability levels: GPT-4o improves from 49\% (FRTR) to 67\%, and Gemini 2.5 Pro reaches 85\% with only 2.23 tool calls. However, smaller models such as LLaMA 4 Maverick 17B and Pixtral Large 2502 show limited gains, suggesting insufficient reasoning capacity for effective multi-step search. In particular, LLaMA 4 Maverick 17B tried to write code instead of calling tools, even though it didn't have access to coding tools. Among baselines, SpreadsheetLLM compression performs poorly on FRTR-Bench (6--34\% accuracy) because SheetCompressor was designed for single-sheet reasoning and compression discards the cross-sheet relationships and multimodal content that FRTR-Bench queries require~\citep{ref_frtr}. Claude in Excel reaches 84\% with Opus 4.6 yet still trails the best BRTR configurations by 15 percentage points, partly because it cannot read embedded images, imposing a ceiling on FRTR-Bench's image-dependent queries. Of FRTR-Bench's 159 queries, 23 require interpreting embedded images; single-pass FRTR achieved only 65\% on these~\citep{ref_frtr}, whereas BRTR's 99\% overall accuracy (with only 2 errors total, both attributable to OCR misreads of embedded image content) demonstrates that iterative image search with context pruning substantially closes this gap.

\paragraph{Results on SpreadsheetLLM Benchmark.}
Table~\ref{tab:spreadsheetllm_results} shows BRTR achieves state-of-the-art performance on the SpreadsheetLLM benchmark, with GPT-5 reaching 98\% accuracy and Claude Sonnet 4.5 achieving 97\%. These results surpass SpreadsheetLLM's own compression approach (89-91\% for frontier models), demonstrating that iterative retrieval generalizes effectively beyond enterprise workbooks to diverse table understanding tasks.

FRTR achieves competitive performance on this benchmark (84--87\% for frontier models), approaching SpreadsheetLLM compression while using comparable token budgets, which suggests single-pass retrieval may suffice for simpler table understanding tasks. Nonetheless, BRTR's 7--11 percentage point improvement over FRTR demonstrates clear value from iterative refinement. The na\"ive full-context baseline significantly underperforms all methods (35--68\% accuracy), confirming that unstructured spreadsheet content overwhelms LLMs without compression or targeted retrieval \citep{ref_lostinthemiddle}.

\paragraph{Discussion.}
Across both benchmarks, BRTR's gains are larger on FRTR-Bench (25 percentage points over single-pass retrieval) than on the SpreadsheetLLM benchmark (7--11 percentage points), reflecting that iterative search most directly addresses the cross-sheet reasoning demands of complex enterprise workbooks. A capability threshold also emerges: frontier models consistently exceed 95\% accuracy with BRTR, while smaller models show a striking reversal on the SpreadsheetLLM benchmark, where they perform worse with BRTR than with SpreadsheetLLM compression. This suggests that effective agentic reasoning requires not only function-calling capabilities but also the capacity to plan multi-step search strategies and refine queries based on partial evidence.

\subsection{Experiment 3: FINCH Finance \& Accounting Benchmark}

To evaluate BRTR's generalization to composite, multi-step enterprise workflows, we assess the framework on FINCH~\citep{ref_finch}, a benchmark of 172 enterprise-grade finance and accounting workflows sourced from authentic enterprise environments including Enron and financial institutions. FINCH workflows operate over 1,710 spreadsheets with 27 million cells, along with PDFs, images, and other artifacts, spanning calculation, structuring, data entry, cross-file retrieval, financial modeling, visualization, and web search tasks. The benchmark captures the compositional and messy nature of real F\&A work: 78.5\% of workflows involve multiple interdependent tasks, and the median workflow covers 15K cells across 8 sheets. The best accuracy reported in the original FINCH paper is 41.9\% (automated evaluation), establishing this as a challenging test of agentic spreadsheet capabilities.

\paragraph{Experimental Setup.}
Using the BRTR planner-executor architecture described in Section~\ref{subsec:planner_executor}, we evaluate five models: Claude Opus 4.6, GPT-5.2, Claude Sonnet 4.5, Gemini 3 Pro, and LLaMA 4 Maverick 17B. Correctness is determined by expert human reviewers following the same evaluation protocol as Experiment~2, with LLM-based pre-screening and final pass/fail decisions made by human evaluators. We compare against a na\"ive baseline of single-call API-based agents evaluated on the same 172 workflows.

\begin{table}[t]
\centering
\caption{Cross-technique performance on the FINCH benchmark (172 workflows). Na\"ive denotes single-call API-based agents; results for Claude Sonnet 4.5 and Gemini 3 Pro are reported by~\citet{ref_finch}. Operational cost columns report medians to account for heavy-tailed workflow distributions. Bold indicates highest accuracy per technique. Models sorted by BRTR accuracy.}
\label{tab:finch_results}
\resizebox{\columnwidth}{!}{%
\begin{tabular}{@{}l cc ccc@{}}
\toprule
& \multicolumn{2}{c}{\textbf{Accuracy}} & \multicolumn{3}{c}{\textbf{BRTR Cost (Median)}} \\
\cmidrule(lr){2-3} \cmidrule(lr){4-6}
Model & BRTR & Na\"ive & Calls & Tok. (K) & Lat. (min) \\
\midrule
Claude Opus 4.6      & \underline{\textbf{0.95}} & \textbf{0.63} & 37 & 358 & 4.8 \\
GPT-5.2              & \underline{\textbf{0.95}}          & 0.46          & 33 & 223 & 4.3 \\
Claude Sonnet 4.5    & 0.87          & 0.20          & 25 & 305 & 2.7 \\
Gemini 3 Pro         & 0.73          & 0.27          & 28 & 380 & 7.9 \\
LLaMA 4 Maverick 17B & 0.12          & 0.01          & 8  & 113 & 0.5 \\
\bottomrule
\end{tabular}%
}
\end{table}

\paragraph{Results.}
Table~\ref{tab:finch_results} presents accuracy on FINCH. BRTR with Claude Opus 4.6 achieves the highest accuracy at 95.3\% (164/172 workflows). GPT-5.2 follows at 94.8\% (163/172). The improvements over na\"ive single-call agents are substantial: +48.9 percentage points for GPT-5.2, +66.9 for Claude Sonnet 4.5, and +46.0 for Gemini~3~Pro. Operational costs are heavily right-skewed, so we report medians; GPT-5.2 achieves the best efficiency trade-off at 223K tokens per workflow compared to 358K for Opus, while trailing by only 0.5 percentage points. LLaMA 4 Maverick 17B achieves only 11.6\%, consistent with the capability threshold observed in Experiment~2.

\paragraph{Discussion.}
BRTR's planner-driven decomposition directly addresses the long-horizon composition bottleneck identified in prior work~\citep{ref_finch}, where accuracy drops sharply on workflows with more than two tasks. By converting composite workflows into a dependency graph with concurrent execution of independent branches, the planner prevents the error accumulation that degrades monolithic approaches. The architecture also mitigates the primary failure modes from prior evaluations: the planner decomposes ambiguous instructions, the search executor enables iterative cross-sheet navigation (Appendix~\ref{app:finch_executors} details per-model executor utilization), formula reasoning benefits from incremental validation within the excel executor, and structured tool calls eliminate code generation errors inherent in monolithic scripts. Of the 8 tasks BRTR fails on the full benchmark, failures cluster into formula evaluation errors (formulas written correctly but not evaluating), instruction-following omissions (missing a sub-step or violating constraints), and incomplete multi-sheet generation, suggesting that execution fidelity rather than retrieval or planning is the primary bottleneck.
\subsection{Ablation Study}

To isolate the contribution of each BRTR component, we evaluate four ablated configurations on a representative 50-task subset of FINCH using Claude Opus 4.6. The subset preserves the full benchmark's distribution across business types and task categories (all within $\pm$4\%), covers all artifact modalities (spreadsheets, PDFs, images, web search), and intentionally includes all six tasks on which BRTR fails on the full benchmark, biasing the subset toward harder cases and yielding a conservative accuracy estimate.

\begin{figure*}[t]
    \centering
    \includegraphics[width=0.72\textwidth]{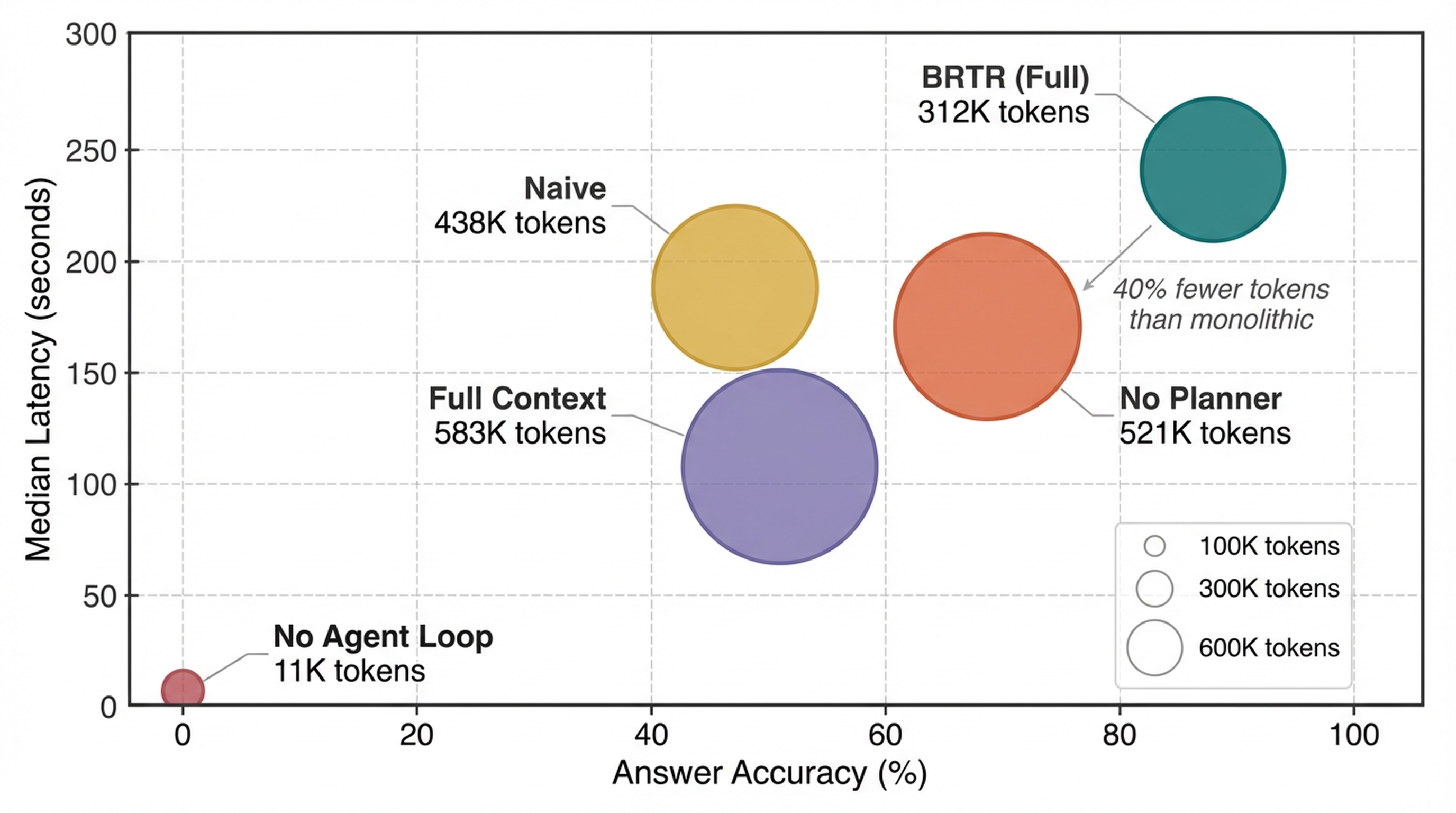}
    \caption{Ablation study on a 50-task FINCH subset (Claude Opus 4.6). Each bubble represents one configuration; position encodes accuracy (x) and latency (y), while bubble area encodes median token consumption. BRTR (Full) achieves the highest accuracy with 40\% fewer tokens than the monolithic agent (No Planner), demonstrating that task decomposition improves both accuracy and efficiency.}
    \label{fig:ablation_bubble}
\end{figure*}

\begin{table}[t]
\centering
\caption{Ablation study on a representative 50-task FINCH subset (Claude Opus 4.6). Each row removes one or more BRTR components. Tool Calls, Tokens, and Latency report medians. Bold indicates highest accuracy.}
\label{tab:ablation}
\resizebox{\columnwidth}{!}{%
\begin{tabular}{@{}lcccc@{}}
\toprule
Configuration & Accuracy & Tool Calls & Tokens (K) & Latency (s) \\
\midrule
BRTR (Full)        & \textbf{0.88} & 33 & 312 & 248 \\
No Planner         & 0.68          & 25 & 521 & 175 \\
Na\"ive            & 0.46          & -- & 438 & 190 \\
Full Context       & 0.52          & 16 & 583 & 114 \\
No Agent Loop      & 0.00          & 1  & 11  & 5 \\
\bottomrule
\end{tabular}%
}
\end{table}

\paragraph{Configurations.}
\emph{No Planner} removes the planner-executor architecture: a single monolithic agent accesses the full tool set across all executors without task decomposition or specialized routing, retaining the iterative agent loop and hybrid BRTR retrieval. \emph{Na\"ive} is the Claude Opus 4.6 API without any BRTR retrieval, as performed in the FINCH evaluation~\citep{ref_finch}. The LLM accesses excel files by reading ranges directly. \emph{Full Context} replaces retrieval with direct file injection, serializing source files into the LLM prompt; the iterative loop and write tools are retained but all search tools are removed. \emph{No Agent Loop} removes the iterative agent loop: a single LLM call with tools bound, where tool calls are executed but outputs are not fed back for further reasoning.

\paragraph{Results.}
Table~\ref{tab:ablation} and Figure~\ref{fig:ablation_bubble} present the ablation results. BRTR achieves 88\% accuracy on the subset, compared to 68\% for No Planner ($-$20 percentage points), 46\% for Na\"ive ($-$42), 52\% for Full Context ($-$36), and 0\% for No Agent Loop. Each component contributes meaningfully, but the planner delivers the largest single improvement (Appendix~\ref{app:ablation_categories} breaks down accuracy by task category).

The planner's impact extends beyond accuracy to token efficiency: despite issuing more tool calls (median 33 vs.\ 25), BRTR consumes 40\% fewer tokens (312K vs.\ 521K) than No Planner. By decomposing tasks into focused sub-tasks with smaller, specialized context windows, the planner avoids the context inflation that occurs when a single agent accumulates outputs from all 43 tools in one growing conversation. Full Context illustrates the cost of bypassing retrieval: it consumes the most tokens (median 583K) yet achieves only 52\% accuracy, because serializing entire files floods the context with irrelevant data. Moreover, 7 of 50 tasks exceed the 200K-token context window when files are injected directly, and 3 web-search tasks have no source files to inject, rendering 20\% of tasks unreachable without retrieval or web tools.

Removing the agent loop entirely (No Agent Loop) results in 0\% accuracy, demonstrating that multi-turn reasoning is essential: a single inference pass produces tool calls that are executed but, without feedback, the model cannot verify results, correct errors, or chain dependent operations.

\subsection{Cost Analysis}

BRTR's iterative approach increases token consumption relative to single-pass baselines. On FRTR-Bench, frontier models consume 20K--92K tokens per query (Table~\ref{tab:frtr_bench_results}), compared to 7.7K for FRTR and 12.7K for SpreadsheetLLM. On FINCH, the most capable configurations consume a median of 223K--358K tokens per workflow (Table~\ref{tab:finch_results}), reflecting the multi-step nature of composite enterprise tasks. GPT-5.2 consistently achieves the best efficiency--accuracy trade-off across all benchmarks, matching frontier accuracy with lower token consumption.

For production deployments, several optimizations can reduce cost. Prompt caching \citep{ref_promptcaching} avoids re-processing the system prompt and tool definitions across turns, which is particularly effective for BRTR's multi-turn conversations where the static prefix dominates early-turn token counts. Semantic caching can further reduce redundant LLM calls for recurring query patterns. As tool registries grow, efficient memory management and dynamic tool selection become critical to preventing context inflation~\citep{ref_memtool, ref_scalemcp, ref_toolshed}.

\section{Conclusion}

We presented Beyond Rows to Reasoning (BRTR), a multimodal agentic framework for spreadsheet understanding that replaces single-pass retrieval with an iterative tool-calling loop, supporting end-to-end Excel workflows from complex analysis to structured editing. A comprehensive evaluation of five multimodal embedding models on FRTR-Bench identified NVIDIA NeMo Retriever 1B as the top performer for mixed tabular and visual data. Across nine LLMs, supported by over 200 hours of expert human evaluation, BRTR achieved state-of-the-art performance across three frontier spreadsheet understanding benchmarks, surpassing prior methods by 25 percentage points on FRTR-Bench, 7 points on SpreadsheetLLM, and 32 points on FINCH. Ablation experiments confirmed that the planner, retrieval, and iterative reasoning each contribute substantially, and cost analysis showed GPT-5.2 achieves the best efficiency--accuracy trade-off. Throughout all evaluations, BRTR maintained full auditability through explicit tool-call traces. Future work will explore extending BRTR to additional structured document types beyond spreadsheets, incorporating learned retrieval strategies that adapt to query complexity, and evaluating the framework in production enterprise deployments with real-time user feedback.

\clearpage

\bibliographystyle{plainnat}
\bibliography{references}

@inproceedings{ref_spreadsheetllm,
  author    = {Dong, Haoyu and Zhao, Jianbo and Tian, Yuzhang and Xiong, Junyu and Xia, Shiyu and Zhou, Mengyu and Lin, Yun and Cambronero, Jos{\'e} and He, Yeye and Han, Shi and Zhang, Dongmei},
  title     = {{SpreadsheetLLM}: Encoding Spreadsheets for Large Language Models},
  booktitle = {Proceedings of the 2024 Conference on Empirical Methods in Natural Language Processing (EMNLP)},
  pages     = {1154--1168},
  year      = {2024},
  address   = {Miami, FL},
}

@misc{ref_frtr,
      title={From Rows to Reasoning: A Retrieval-Augmented Multimodal Framework for Spreadsheet Understanding}, 
      author={Anmol Gulati and Sahil Sen and Waqar Sarguroh and Kevin Paul},
      year={2026},
      eprint={2601.08741},
      archivePrefix={arXiv},
      primaryClass={cs.CL},
      url={https://arxiv.org/abs/2601.08741}, 
}

@inproceedings{ref_tapas,
  author    = {Herzig, Jonathan and Nowak, Pawe{\l} Krzysztof and M{\"u}ller, Thomas and Piccinno, Francesco and Eisenschlos, Julian},
  title     = {{TaPas}: Weakly Supervised Table Parsing via Pre-training},
  booktitle = {Proceedings of the 58th Annual Meeting of the Association for Computational Linguistics (ACL)},
  pages     = {4320--4333},
  year      = {2020},
}

@inproceedings{ref_tuta,
  author    = {Wang, Zhiruo and Dong, Haoyu and Jia, Ran and Li, Jia and Fu, Zhiyi and Han, Shi and Zhang, Dongmei},
  title     = {{TUTA}: Tree-based Transformers for Generally Structured Table Pre-training},
  booktitle = {Proceedings of the 27th ACM SIGKDD Conference on Knowledge Discovery and Data Mining (KDD)},
  pages     = {1780--1790},
  year      = {2021},
}

@inproceedings{ref_sheetagent,
  author    = {Chen, Yibin and others},
  title     = {{SheetAgent}: A Generalist Agent for Spreadsheet Reasoning and Manipulation via Large Language Models},
  booktitle = {Proceedings of the ACM Web Conference 2025 (WWW)},
  year      = {2025},
}

@article{ref_tablezoomer,
  author    = {Xiong, Sishi and He, Ziyang and He, Zhongjiang and Zhao, Yu and Pan, Changzai and Zhang, Jie and Wu, Zhenhe and Song, Shuangyong and Li, Yongxiang},
  title     = {{TableZoomer}: A Collaborative Agent Framework for Large-scale Table Question Answering},
  journal   = {arXiv preprint arXiv:2509.01312},
  year      = {2025},
}

@inproceedings{ref_react,
  author    = {Yao, Shunyu and Zhao, Jeffrey and Yu, Dian and Du, Nan and Shafran, Izhak and Narasimhan, Karthik and Cao, Yuan},
  title     = {{ReAct}: Synergizing Reasoning and Acting in Language Models},
  booktitle = {Proceedings of the International Conference on Learning Representations (ICLR)},
  year      = {2023},
}

@inproceedings{ref_tapex,
  author    = {Liu, Qian and others},
  title     = {{TAPEX}: Table Pre-training via Learning a Neural {SQL} Executor},
  booktitle = {Proceedings of the International Conference on Learning Representations (ICLR)},
  year      = {2022},
}

@inproceedings{ref_fortap,
  author    = {Cheng, Zhoujun and Dong, Haoyu and Jia, Ran and Wu, Pengfei and Han, Shi and Cheng, Fan and Zhang, Dongmei},
  title     = {{ForTaP}: Using Formulas for Numerical-Reasoning-Aware Table Pretraining},
  booktitle = {Proceedings of the 60th Annual Meeting of the Association for Computational Linguistics (ACL)},
  pages     = {1150--1166},
  year      = {2022},
  address   = {Dublin, Ireland},
}

@article{ref_fortune,
  author    = {Wang, Yilun and others},
  title     = {Fortune: Formula-Driven Reinforcement Learning for Symbolic Table Reasoning in Language Models},
  journal   = {arXiv preprint arXiv:2505.23667},
  year      = {2025},
}

@article{ref_finch,
  author    = {Dong, Haoyu and Zhang, Pengkun and Gao, Yan and Dong, Xuanyu and Cheng, Yilin and Lu, Mingzhe and Yakefu, Adina and Zheng, Shuxin},
  title     = {{FINCH}: Benchmarking Finance \& Accounting across Spreadsheet-Centric Enterprise Workflows},
  journal   = {arXiv preprint arXiv:2512.13168},
  year      = {2025},
}

@article{ref_sheetbrain,
  author    = {Wang, Zhonghao and others},
  title     = {{SheetBrain}: A Neuro-Symbolic Agent for Accurate Reasoning over Complex and Large Spreadsheets},
  journal   = {arXiv preprint arXiv:2510.19247},
  year      = {2025},
}

@article{ref_sheetmind,
  author    = {Zhu, Rongyu and Cheng, Xiang and Liu, Kevin and Zhu, Bowei and Jin, Di and Parihar, Nikhil and Xu, Ziming and Gao, Oliver},
  title     = {{SheetMind}: An End-to-End {LLM}-Powered Multi-Agent Framework for Spreadsheet Automation},
  journal   = {arXiv preprint arXiv:2506.12339},
  year      = {2025},
}

@article{ref_tablemind,
  author    = {Jiang, Cong and Cheng, Mingda and Tao, Xin and Mao, Qinglong and Ouyang, Jun and Liu, Qi},
  title     = {{TableMind}: An Autonomous Programmatic Agent for Tool-Augmented Table Reasoning},
  journal   = {arXiv preprint arXiv:2509.06278},
  year      = {2025},
}

@misc{ref_llamasheets,
  author    = {{LlamaIndex}},
  title     = {{LlamaSheets}: Turn Messy Spreadsheets into {AI}-Ready Data},
  howpublished = {\url{https://www.llamaindex.ai/}},
  year      = {2025},
}

@misc{ref_claudeexcel,
  author    = {{Anthropic}},
  title     = {Claude in {Excel}: {AI}-Powered Spreadsheet Assistant},
  howpublished = {\url{https://support.anthropic.com/en/articles/12650343-claude-in-excel}},
  year      = {2025},
}

@misc{ref_copilotagent,
  author    = {{Microsoft}},
  title     = {Building Agent Mode in {Excel}},
  howpublished = {\url{https://techcommunity.microsoft.com/blog/excelblog/building-agent-mode-in-excel/4457320}},
  year      = {2025},
}

@misc{ref_geminisheets,
  author    = {{Google}},
  title     = {Collaborate with {Gemini} in {Google Sheets}},
  howpublished = {\url{https://support.google.com/docs/answer/14356410}},
  year      = {2025},
}

@misc{ref_claudecowork,
  author    = {{Anthropic}},
  title     = {Claude Cowork: Agentic Desktop {AI} for Knowledge Work},
  howpublished = {\url{https://support.claude.com/en/articles/13345190-getting-started-with-cowork}},
  year      = {2026},
}

@inproceedings{ref_spreadsheetbench,
  author    = {Ma, Zeyao and others},
  title     = {{SpreadsheetBench}: Towards Challenging Real World Spreadsheet Manipulation},
  booktitle = {Advances in Neural Information Processing Systems (NeurIPS)},
  year      = {2024},
}

@article{ref_agenticrag,
  author    = {Singh, Aditi and Ehtesham, Abul and Kumar, Saket and Talaei Khoei, Tala},
  title     = {Agentic Retrieval-Augmented Generation: A Survey on Agentic {RAG}},
  journal   = {arXiv preprint arXiv:2501.09136},
  year      = {2025},
}

@inproceedings{ref_rrf,
  author    = {Cormack, Gordon V. and Clarke, Charles L. A. and Buettcher, Stefan},
  title     = {Reciprocal Rank Fusion Outperforms Condorcet and Individual Rank Learning Methods},
  booktitle = {Proceedings of the 32nd International ACM SIGIR Conference on Research and Development in Information Retrieval (SIGIR)},
  pages     = {758--759},
  year      = {2009},
}

@article{ref_rag,
  author    = {Lewis, Patrick and Perez, Ethan and Piktus, Aleksandra and Petroni, Fabio and Karpukhin, Vladimir and Goyal, Naman and K{\"u}ttler, Heinrich and Lewis, Mike and Yih, Wen-tau and Rockt{\"a}schel, Tim and Riedel, Sebastian and Kiela, Douwe},
  title     = {Retrieval-Augmented Generation for Knowledge-Intensive {NLP} Tasks},
  journal   = {arXiv preprint arXiv:2005.11401},
  year      = {2021},
}

@article{ref_lostinthemiddle,
  author    = {Liu, Nelson F. and Lin, Kevin and Hewitt, John and Paranjape, Ashwin and Bevilacqua, Michele and Petroni, Fabio and Liang, Percy},
  title     = {Lost in the Middle: How Language Models Use Long Contexts},
  journal   = {arXiv preprint arXiv:2307.03172},
  year      = {2023},
}

@article{ref_instructexcel,
  author    = {Payan, Justin and Mishra, Swaroop and Singh, Mukul and Negreanu, Carina and Poelitz, Christian and Baral, Chitta and Roy, Subhro and Chakravarthy, Rasika and Van Durme, Benjamin and Nouri, Elnaz},
  title     = {{InstructExcel}: A Benchmark for Natural Language Instruction in {Excel}},
  journal   = {arXiv preprint arXiv:2310.14495},
  year      = {2023},
}

@article{ref_promptcaching,
  author    = {Lumer, Elias and Nizar, Faheem and Jangiti, Akshaya and Frank, Kevin and Gulati, Anmol and Phadate, Mandar and Subbiah, Vamse Kumar},
  title     = {Don't Break the Cache: An Evaluation of Prompt Caching for Long-Horizon Agentic Tasks},
  journal   = {arXiv preprint arXiv:2601.06007},
  year      = {2026},
}

@article{ref_memtool,
  author    = {Lumer, Elias and Gulati, Anmol and Subbiah, Vamse Kumar and Basavaraju, Pradeep Honaganahalli and Burke, James A.},
  title     = {{MemTool}: Optimizing Short-Term Memory Management for Dynamic Tool Calling in {LLM} Agent Multi-Turn Conversations},
  journal   = {arXiv preprint arXiv:2507.21428},
  year      = {2025},
}

@article{ref_scalemcp,
  author    = {Lumer, Elias and Gulati, Anmol and Subbiah, Vamse Kumar and Basavaraju, Pradeep Honaganahalli and Burke, James A.},
  title     = {{ScaleMCP}: Dynamic and Auto-Synchronizing Model Context Protocol Tools for {LLM} Agents},
  journal   = {arXiv preprint arXiv:2505.06416},
  year      = {2025},
}

@article{ref_toolshed,
  author    = {Lumer, Elias and Subbiah, Vamse Kumar and Burke, James A. and Basavaraju, Pradeep Honaganahalli and Huber, Austin},
  title     = {Toolshed: Scale Tool-Equipped Agents with Advanced {RAG}-Tool Fusion and Tool Knowledge Bases},
  journal   = {arXiv preprint arXiv:2410.14594},
  year      = {2024},
}

@article{ref_toolagentselection,
  author    = {Lumer, Elias and Gulati, Anmol and Nizar, Faheem and Hedroits, Dilan and Mehta, Ankit and Hwangbo, Hyunji and Subbiah, Vamse Kumar and Basavaraju, Pradeep Honaganahalli and Burke, James A.},
  title     = {Tool and Agent Selection for Large Language Model Agents in Production: A Survey},
  journal   = {Preprints},
  year      = {2025},
  doi       = {10.20944/preprints202512.1050.v1},
}

@article{ref_spreadsheetarena,
  author    = {Kundurthy, Srivatsa and Na, Clara and Handley, Michael and Kirshner, Zach and Zhang, Chen Bo Calvin and Sharma, Manasi and Strubell, Emma and Ling, John},
  title     = {{SpreadsheetArena}: Decomposing Preference in {LLM} Generation of Spreadsheet Workbooks},
  year      = {2026},
  note      = {Preprint, February 2026},
  url       = {https://www.meridian.ai/research/SpreadsheetArena.pdf},
}

\clearpage
\onecolumn
\appendix

\section{FINCH Per-Model Executor Utilization}\label{app:finch_executors}

Table~\ref{tab:finch_executors} reports executor-level statistics for each model on the full FINCH benchmark. The planner dispatches sub-tasks to specialized executors; invocation counts reflect how many sub-tasks each executor handles. The excel executor dominates across all models, consistent with FINCH being spreadsheet-centric. Notably, Claude Opus 4.6 invokes the search executor on 148 of 172 tasks, suggesting aggressive use of retrieval, while Claude Sonnet 4.5 uses search on only 87 tasks yet still achieves 87.2\% accuracy. LLaMA 4 Maverick 17B lacks executor statistics because it failed to produce valid planner outputs for most tasks.

\begin{table}[h]
\centering
\caption{FINCH executor utilization per model. Invocations counts sub-tasks dispatched to each executor; Tasks counts distinct workflows using that executor; Tool Calls counts individual tool invocations within the executor. LLaMA 4 Maverick 17B omitted due to insufficient valid planner outputs.}
\label{tab:finch_executors}
\begin{tabular}{@{}l l ccc@{}}
\toprule
Model (Accuracy) & Executor & Invocations & Tasks & Tool Calls \\
\midrule
Claude Opus 4.6 (95.3\%) & Search & 264 & 148 & 1{,}623 \\
 & Excel & 232 & 166 & 5{,}351 \\
 & IO & 16 & 14 & 271 \\
 & Web & 18 & 11 & 651 \\
\midrule
GPT-5.2 (94.8\%) & Search & 197 & 145 & 1{,}164 \\
 & Excel & 425 & 164 & 5{,}330 \\
 & IO & 25 & 14 & 142 \\
 & Web & 12 & 11 & 272 \\
\midrule
Claude Sonnet 4.5 (87.2\%) & Search & 110 & 87 & 831 \\
 & Excel & 248 & 162 & 4{,}179 \\
 & IO & 15 & 7 & 80 \\
 & Web & 11 & 11 & 574 \\
\midrule
Gemini 3 Pro (73.3\%) & Search & 152 & 148 & 715 \\
 & Excel & 342 & 165 & 5{,}225 \\
 & IO & 25 & 17 & 237 \\
 & Web & 11 & 11 & 380 \\
\bottomrule
\end{tabular}
\end{table}

\section{Ablation Accuracy by Task Category}\label{app:ablation_categories}

Table~\ref{tab:ablation_categories} disaggregates ablation results by task category on the 50-task FINCH subset. Each task may belong to multiple categories. The planner contributes most on cross-sheet/file retrieval (86\% vs.\ 43\% for No Planner), where task decomposition enables the search executor to navigate across files. Retrieval is critical for web search (67\% vs.\ 0\% for Full Context) and translation tasks (100\% vs.\ 0\%), which have no source files to inject. Calculation tasks show the smallest planner gap (94\% vs.\ 69\%), suggesting that formula reasoning is less dependent on task decomposition.

\begin{table}[h]
\centering
\caption{Ablation accuracy by task category on the 50-task FINCH subset (Claude Opus 4.6). Each task may span multiple categories. $n$ denotes the number of tasks involving each category.}
\label{tab:ablation_categories}
\begin{tabular}{@{}lcccc@{}}
\toprule
Category ($n$) & BRTR & $-$ Planner & $-$ Retrieval & $-$ Iteration \\
\midrule
Calculation (35)             & 0.94 & 0.69 & 0.54 & 0.00 \\
Structuring / Formatting (23) & 0.87 & 0.74 & 0.48 & 0.00 \\
Cross-sheet Retrieval (14)   & 0.86 & 0.43 & 0.43 & 0.00 \\
Validation / Review (11)     & 0.82 & 0.64 & 0.45 & 0.00 \\
Data Entry / Import (11)     & 0.91 & 0.64 & 0.45 & 0.00 \\
Summary / Visualization (10) & 0.90 & 0.70 & 0.50 & 0.00 \\
Financial Modeling (4)       & 0.75 & 1.00 & 0.50 & 0.00 \\
Web Search (3)               & 0.67 & 0.33 & 0.00 & 0.00 \\
Translation (2)              & 1.00 & 0.50 & 0.00 & 0.00 \\
\bottomrule
\end{tabular}
\end{table}

Failure overlap analysis reveals that the planner and retrieval components rescue largely distinct task sets: the planner rescues 12 tasks that fail under No Planner, while retrieval rescues 21 tasks that fail under Full Context. Only 3 tasks (IDs 9, 2, 17) fail across all configurations, indicating hard cases that neither decomposition nor retrieval alone can resolve. Two tasks (IDs 24, 31) fail with BRTR but pass without the planner, suggesting that over-decomposition occasionally harms performance on tasks better handled monolithically.

\section{Example Tool-Call Trace}\label{app:tool_trace}

\begin{figure}[h]
\centering
\includegraphics[width=\textwidth]{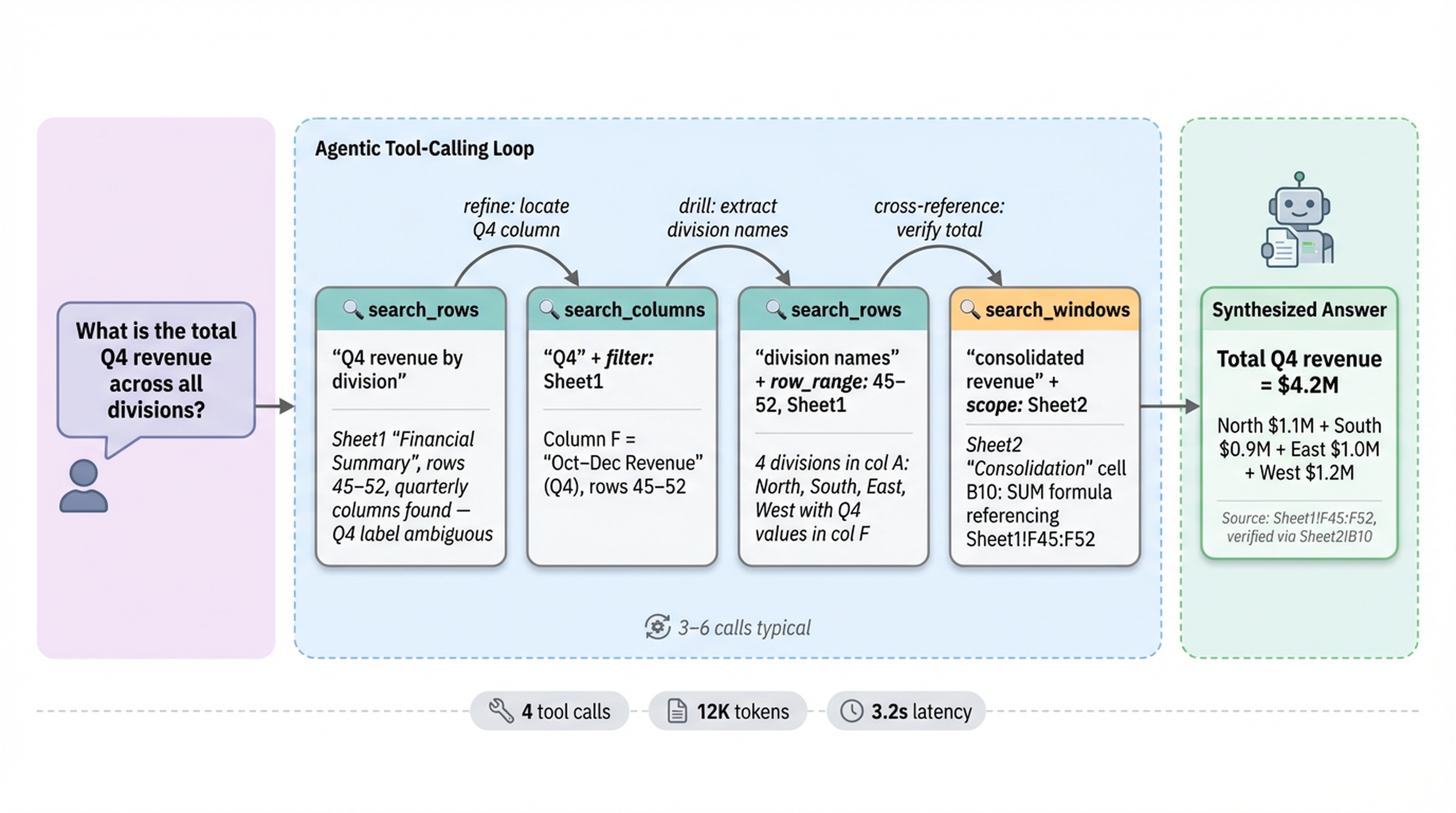}
\caption{Example BRTR tool-call trace for a representative query. The agent iteratively invokes search tools, inspects returned chunks, refines queries, and cross-references sheets before synthesizing a grounded answer with full provenance.}
\label{fig:tool_trace}
\end{figure}

Figure~\ref{fig:tool_trace} illustrates a representative BRTR tool-call trace on an FRTR-Bench query over a multi-sheet enterprise workbook. The agent begins with a broad row search, then iteratively refines by locating the target column, extracting division-level values, and cross-referencing a consolidation sheet before synthesizing a fully grounded answer with cell-level provenance. This four-call trace is typical of frontier model behavior (3--6 calls on average) and demonstrates how each intermediate result informs the next query, a capability absent in single-pass pipelines.

\section{Planner Decomposition Prompt}\label{app:decomposition_prompt}

The following is a condensed version of the prompt template used by the BRTR planner to decompose composite workflows into sub-task dependency graphs (Phase~1 of Algorithm~\ref{alg:brtr_planner}). Before invoking the LLM, the planner searches the vector index for column locations matching quoted terms in the task instruction; results are injected via the \texttt{\{exploration\_context\}} placeholder to ground decomposition in the actual data schema.

\begin{small}
\begin{verbatim}
Analyze this task and return a JSON plan.
Task: {task}
Output path: {output_path}
{exploration_context}

EXECUTOR TYPES:
- "search": Query indexed Excel files (column locations, values, structure)
- "excel": Create/modify workbooks, write data/formulas, format cells.
           Has OCR via transcribe_image tool for reading images/PDFs.
- "io": Read/write non-Excel files (CSV, JSON, DOCX, PDF, Markdown).
        Use write_pdf for PDF output, write_docx for Word output.
- "web": Fetch external data from the internet when source files
         don't contain needed information.

OUTPUT TYPES: "spreadsheet" | "text" | "document" | "both"

CRITICAL RULES:
1. Create 1-6 subtasks based on task complexity.
2. If the task mentions source file paths, use the EXACT FULL paths.
3. Include output path in subtask descriptions where files are created.
4. If DATA EXPLORATION RESULTS are provided, use the SPECIFIC column letters
   and sheet names discovered.
5. Include ALL filter constraints in EACH subtask (date ranges, thresholds).
6. CROSS-SHEET RETRIEVAL: Search must identify source AND target locations.
7. REPETITIVE TASKS: Split into separate subtasks per section.

Return JSON:
{"output_type": "...", "subtasks": [
  {"id": 1, "type": "search|excel|io|web",
   "description": "...", "dependencies": []}, ...]}

Return JSON ONLY. Maximum 6 subtasks.
\end{verbatim}
\end{small}

\end{document}